\documentclass[conference]{IEEEtran}
\IEEEoverridecommandlockouts
\usepackage{cite}
\usepackage{amsmath,amssymb,amsfonts}
\usepackage{algorithmic}
\usepackage{graphicx}
\usepackage{textcomp}
\usepackage{xcolor}
\usepackage{url}
\usepackage{verbatim}
\usepackage{hyperref}
\usepackage{soul}
\usepackage{graphicx}
\usepackage{float}
\def\BibTeX{{\rm B\kern-.05em{\sc i\kern-.025em b}\kern-.08em
    T\kern-.1667em\lower.7ex\hbox{E}\kern-.125emX}}

\begin{document}


\title{
FOD-A: A Dataset for Foreign \\ Object Debris in Airports

\thanks{
This paper has been accepted for publication by 20th IEEE International Conference on Machine Learning and Applications. The copyright is with the IEEE.
}
}

\author{
\IEEEauthorblockN{Travis Munyer}
\IEEEauthorblockA{\textit{Dept. of Computer Science} \\
\textit{Univ. of Nebraska Omaha}\\
tmunyer@unomaha.edu}

\and

\IEEEauthorblockN{Pei-Chi Huang}
\IEEEauthorblockA{\textit{Dept. of Computer Science} \\
\textit{Univ. of Nebraska Omaha}\\
phuang@unomaha.edu}

\and

\IEEEauthorblockN{Chenyu Huang}
\IEEEauthorblockA{\textit{Aviation Institute} \\
\textit{Univ. of Nebraska Omaha}\\
chenyuhuang@unomaha.edu}

\and

\IEEEauthorblockN{Xin Zhong}
\IEEEauthorblockA{\textit{Dept. of Computer Science} \\
\textit{Univ. of Nebraska Omaha}\\
xzhong@unomaha.edu}

}

\maketitle

\begin{abstract}
Foreign Object Debris (FOD) detection has attracted increased attention in the area of machine learning and computer vision. However, a robust and publicly available image dataset for FOD has not been initialized. To this end, this paper introduces an image dataset of FOD, named FOD in Airports (FOD-A). FOD-A object categories have been selected based on guidance from prior documentation and related research by the Federal Aviation Administration (FAA). In addition to the primary annotations of bounding boxes for object detection, FOD-A provides labeled environmental conditions. As such, each annotation instance is further categorized into three light level categories (\textit{bright, dim,} and \textit{dark}) and two weather categories (\textit{dry} and \textit{wet}). Currently, FOD-A has released 31 object categories and over 30,000 annotation instances.  This paper presents the creation methodology, discusses the publicly available dataset extension process, and demonstrates the practicality of FOD-A with widely used machine learning models for object detection. 
\end{abstract}


\section{Introduction}\label{sec:intro}
\label{intro}
Accidents caused by Foreign Object Debris (FOD) are responsible for severe injuries or death, and billions of dollars in damages to aircraft~\cite{faa-fact-sheet}. FOD is a critical safety hazard in airports, and machine learning and computer vision (MLCV) technology has been shown as a potential solution in exploratory research~\cite{Yuan_2020, YOLOV3forFOD, FOD-conv-nn, cnn-for-FOD}. In order to facilitate this application of MLCV, a dataset of FOD images and annotations is required to be utilized and organized for more sophisticated and robust models and algorithms.

\begin{figure}[t]
    \centering
    \includegraphics[width=0.4\textwidth]{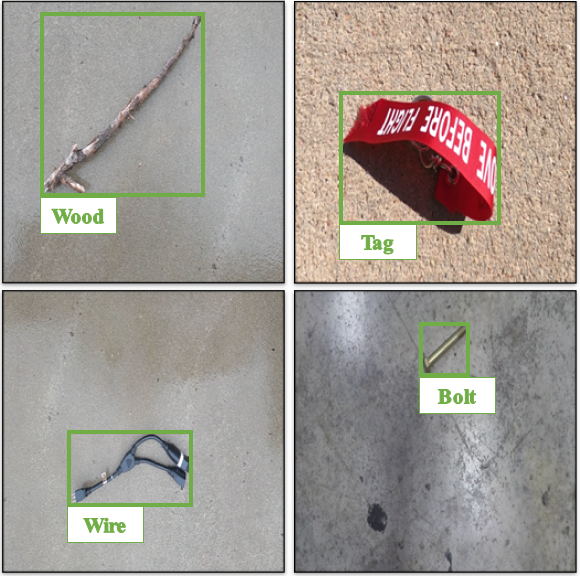}
    \vspace{-.75em}
    \caption{FOD-A dataset images with bounding box annotation examples.}
    \vspace{-2em}
    \label{fig:dataset_example}
\end{figure}

To foster future FOD-based MLCV work, we developed the novel dataset named Foreign Object Debris in Airports (FOD-A). FOD-A object categories are influenced by relevant FAA documents and previous research~\cite{fod-management, FOD-detection-equipment, FOD-Finder-assessment, Tarsier-radar-assessment, iFerret_assessment, FODetect-Assessment, faa-fact-sheet}. These object categories are designed to cover several FOD types while including specific labels that are descriptive. Images are collected under varying light and weather conditions to ensure accurate simulation of typical airport environments. Varying conditions also ensure that the FOD-A dataset is challenging for modern MLCV algorithms. As a preview, FOD-A images with example bounding boxes are shown in Figure~\ref{fig:dataset_example}. Since FOD is a continually evolving datatype, it is important that a FOD dataset can be easily expanded. To enable extensibility, this dataset includes tools that allow the addition of new data with ease. Since extensibility and consistency may be contradictory traits, it will be important for several iterations of FOD-A to remain available. Once algorithms are compared using a consistent iteration of FOD-A, new algorithms can be implemented using the most current FOD-A data. This ensures all current FOD-A object categories are included in final detection algorithms.

Currently, there are several well-known general datasets available that contain diverse categories of common objects, (\textit{e.g}., bicycles, cars, desks, toasters)~\cite{Everingham15, lin2014microsoft}. Due to the location of the FOD datatype (\textit{i.e.} airports), these datasets~\cite{Everingham15, lin2014microsoft} do not properly cover necessary categories of FOD (\textit{e.g}., luggage items, aircraft parts, tools). Because of this, comparisons to datasets of general objects will only be briefly made in Section~\ref{dataset-stats}. 

The  rest  of  this  paper  is  organized  as  follows. The related work is reviewed in Section~\ref{sec:related}. A description of dataset creation methodology, FOD-A statistics, and the extension process is provided in Section~\ref{sec:construction}. Section~\ref{sec:analysis} presents the initial experimentation and algorithmic results. Finally, Section~\ref{sec:conclusion_discussion} concludes the paper and discusses future work. As a summary, the major contributions of this work are as follows: 1) the initialization and creation of the novel and publicly available dataset FOD-A; 2) the design and development of an efficient and abstractable method of image dataset creation; and 3) the implementation and evaluation of an initial algorithmic analysis of FOD-A.

\section{Related Work}\label{sec:related}
The FAA published several documents~\cite{faa-fact-sheet, fod-management, FOD-detection-equipment, iFerret_assessment, FODetect-Assessment, FOD-Finder-assessment, Tarsier-radar-assessment} providing guidance on FOD detection and management. As the main resource for FOD information, the object categories in the FOD dataset are based off the FAA documentation~\cite{fod-management, FOD-detection-equipment}. The details of the category selection process are described in Section~\ref{image-collection}.

\subsection{Existing FOD Datasets}
A publicly accessible FOD dataset~\cite{FOD-conv-nn} does exist. However, this dataset primarily focused on material recognition, including the following three object categories: metal, plastic, and concrete~\cite{FOD-conv-nn}. Using only these three object categories does not cover all common types of FOD according to the FAA's~\cite{FOD-detection-equipment} information. For example, these object categories cannot cover some types of tools, various common airport garbage (\textit{e.g.,} paper, soda cans), animals and other natural debris, some runway materials such as paint chips, and other common FOD~\cite{FOD-detection-equipment, faa-fact-sheet}. For comparison, FOD-A provides 31 object categories (see Figure~\ref{fig:annotation_instances}).

Furthermore, images contained in the material recognition dataset are in a zoomed-in format. It is likely that images collected during applied FOD detection tasks will not be zoomed into objects, so FOD-A provides images in a zoomed-out format with bounding boxes (see Figure~\ref{fig:dataset_example}). Also, the material recognition dataset contains about 3000 object instances, while the FOD-A dataset contains over 30,000 object instances. In summary, the FOD-A dataset is more appropriate for FOD detection tasks because it contains forms of annotation better suited to the airport environment (\textit{i.e.} bounding box annotation plus weather and light categorization annotation), several more object instances, and descriptive object categories.

\subsection{Related MLCV FOD Detection Methods}

FOD-focused MLCV research is becoming increasingly common~\cite{YOLOV3forFOD, cnn-for-FOD, FOD-conv-nn, Yuan_2020}. Several papers have been published that implement algorithms to detect FOD. These papers have created their own private datasets and have generally kept these datasets small. The effectiveness of major datasets in several research tasks~\cite{lin2014microsoft, Everingham15} can partially be attributed to focused dataset development and publicly accessible versions. It is a much larger task to create a robust dataset~\cite{8614069} when also presenting new detection methods. Thus, FOD-A should enable researchers to focus on the improvement of FOD detection algorithms.

P. Li and H. Li~\cite{YOLOV3forFOD} did create their own small dataset for internal use. Their dataset consists of about 2000 images, with 100 images per object class~\cite{YOLOV3forFOD}. Although their small and private dataset may be feasible for their proposed algorithms, it may not be suited to larger scale experiments. 

Having considered the shortcomings and benefits of prior works discussed above, we have developed FOD-A with three main advantages:
1) FOD-A provides a wide range of descriptive object categories selected by FAA documentation, and includes a large number of instances for each category;
2) FOD-A is publicly available with documented expansion processes; 
and 3) FOD-A considers realistic and challenging data samples in varying weather and light conditions.

\section{Dataset Construction}\label{sec:construction}
This section presents FOD-A in more details, including the creation methodology, FOD-A statistics, and the extension process.

\subsection{Image Collection}
\label{image-collection}
According to the FAA, FOD commonly includes the following: ``aircraft and engine fasteners (nuts, bolts, washers, safety wire, etc.); aircraft parts (fuel caps, landing gear fragments, oil sticks, metal sheets, trapdoors, and tire fragments); mechanics' tools; catering supplies; flight line items (nails, personnel badges, pens, pencils, luggage tags, soda cans, etc.); apron items (paper and plastic debris from catering and freight pallets, luggage parts, and debris from ramp equipment); runway and taxiway materials (concrete and asphalt chunks, rubber joint materials, and paint chips); construction debris (pieces of wood, stones, fasteners and miscellaneous metal objects); plastic and/or polyethylene materials; natural materials (plant fragments, wildlife and volcanic ash); and contaminants from winter conditions (snow, ice)'' \cite{fod-management}.

\begin{figure*}[t]
    \centering
    \includegraphics[width=475pt]{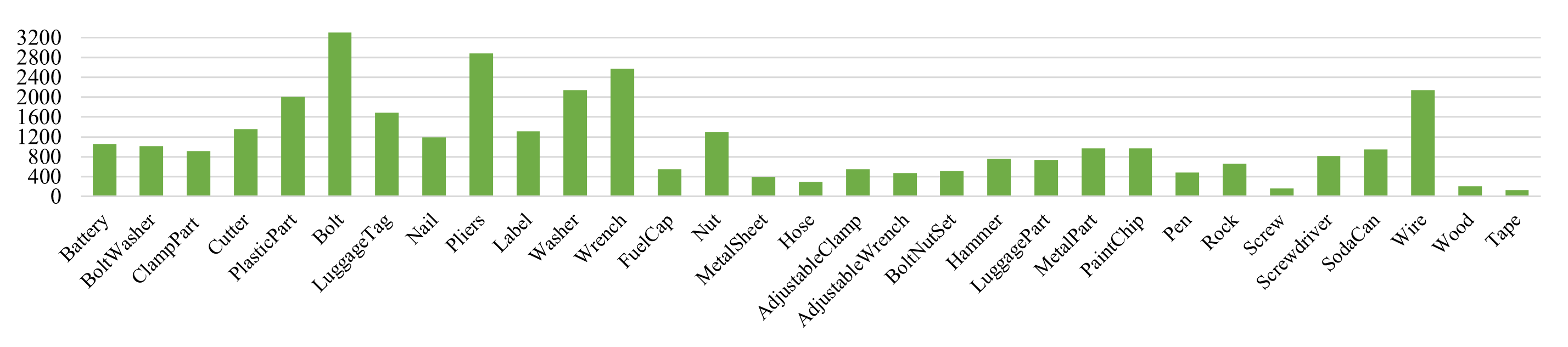}
    \vspace{-1.4em}
    \caption{Instances per category in the FOD-A dataset.}
    \vspace{-1.7em}
    \label{fig:annotation_instances}
\end{figure*}

The FAA suggests that metal objects are the most common FOD (over 60 percent of the materials found in a one-year airport study) \cite{fod-management}. We generalize the FAA description into categories suitable for MLCV applications with priority given to objects that have the highest potential to harm aircraft (\textit{i.e.} become ingested and damage jet engines, shred tires, etc.). The resultant categories for this implementation of the FOD-A dataset can be found in Figure~\ref{fig:annotation_instances}.

\begin{figure}[t]
    \centering
    \includegraphics[width=2.7cm]{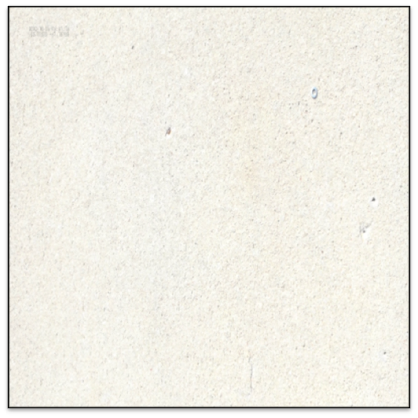}
    \vspace{1mm}
    \includegraphics[width=2.7cm]{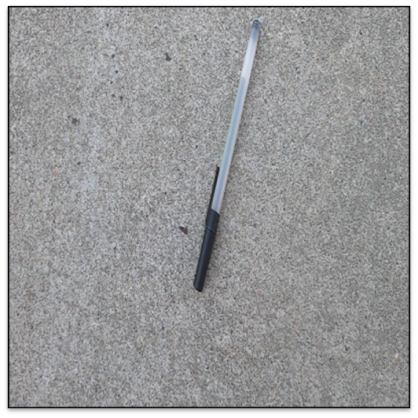}
    \includegraphics[width=2.7cm]{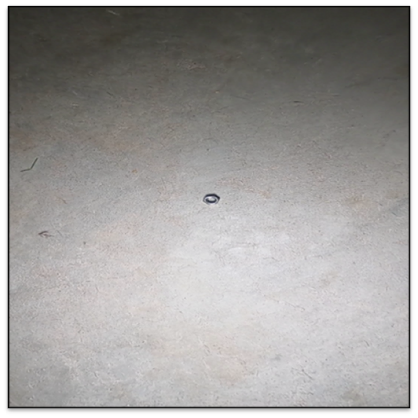}
    \vspace{-1em}
    \caption{Example images from various FOD-A light-level categories. Example \textit{bright} image (left), an example \textit{dim} image (middle), and an example \textit{dark} image (right).}
    \vspace{-1.75em}
    \label{fig:brightness-examples}
\end{figure}

In order to create a practical dataset that is applicable to airport FOD management, we collect images in diverse conditions. Weather and light conditions in airports vary, so a dataset of FOD objects must incorporate this fact into included data. Wet and dry environments provide weather variation for FOD-A image collection. For light variation, the image collection process incorporates bright, dim, and dark light conditions. Since each of these environmental variations could be easily abstracted to fit categorization tasks, FOD-A includes categorization labels for weather (dry and wet) and light-level (bright, dim, and dark). Example light-level categorization images are provided in Figure~\ref{fig:brightness-examples}. Since snow is promptly cleared from the airport environment, it is unnecessary to include a snowy category. Any moisture that remains after snow is cleared should still fit into the wet category. FOD-A's dry and wet weather categories should cover the majority of weather types applicable to airports. Remarkably, the weather and light-level catgorization annotations are in addition to the focus of the FOD-A, which is bounding box annotations for object detection.

Images of common FOD are collected in the video (mp4) format using both portable and unmanned aerial vehicle (UAV) cameras. UAV image collection allows variation in recording distances that could not be achieved with handheld portable cameras. The images gathered by the portable camera were closer to the object, and camera angles changed more drastically than the UAV camera. As a video is densely packed with images, the video format allows for large-scale image collection. However, utilizing the video format presented a few initial issues. Some videos do not have the target object(s) in each frame, and the empty frames could uselessly pollute the dataset. A video would have to be trimmed to proper intervals, and then each frame separated to an image format. Performing this task by hand is very time-consuming and prevents extensibility. We created a small command-line tool to solve a few of these issues. The benefits of this tool are as follows: 1) it allows for that dataset to be easily expanded using simple instructions; 2) it makes the image collection process more efficient; and 3) it normalizes the images.

This tool takes a video file as an input, and then creates a folder structure, a trimmed version of the video, and the frames for the video. The trimmed video can then be used as input into Computer Vision Annotation Tool (CVAT)~\cite{CVAT}, which makes the annotation processes efficient. Further details of the annotation process are described in section~\ref{image-annotation}. Each folder structure, once the annotations are added, corresponds to the set of annotations and frames for one video. The folder structures are stored together; this forms the original FOD-A format. We provide the tools that can convert the FOD-A format to the Pascal Visual Object Classes format (called Pascal VOC)~\cite{Everingham15} as commonly required by algorithmic processes.

The expansion tool is designed to further edit the video by automatically reducing the FPS of each video to 15. The tool then generates the frames after the FPS reduction. Because the original FPS of each video was about $60$ FPS, this is an effective method to prevent duplicate images. The output location for the folders and frames is stored in a settings file to allow efficient expansion. Based on different settings, the expansion tool can automatically apply to either create a new dataset (if targeting a new folder) or to expand an existing dataset (if targeting an existing directory). The abstract design of the tools can enable the process and format to fit any image datatype.

Because each of the original videos include consistent light and weather conditions, the expansion tool automatically generates the categorization annotations once the weather and light conditions are specified. As the tool outputs each individual frame to the folder structure, it saves the correct categorization annotations along with a relative file path to the new image in a Comma-Separated Values (CSV) file.

The produced images and annotations for FOD-A have been uploaded to the the \href{https://github.com/FOD-UNOmaha/FOD-data}{GitHub repository} in the original format and in the Pascal VOC format~\cite{Everingham15}. This GitHub page also contains the detailed dataset expansion instructions. These instructions can also be used to create new image datasets. Additional images can be added to the dataset by inputting more videos into the expansion tool. Required changes to the dataset, such as format extension and data preparation, are made automatically by the expansion tool after additions. Inserting annotations to the created folder structure is the only additional requirement for extension. This enables the dataset expansion process to be efficient.

\subsection{Image Annotation}\label{image-annotation}
To create a robust dataset, it is best to include as many instances of FOD objects as possible. Therefore, an efficient and quality annotation process was implemented. Video data can be quickly and accurately annotated using the existing tracking algorithms for videos. The efficient annotation process was provided by the open source tool, CVAT~\cite{CVAT}. Since the FOD-A expansion tool outputs a matching trimmed video as well as the video's frames, the images are annotated as if they are still a video format. CVAT's video annotation algorithm only requires every ten frames (an interval that can be modified) to be annotated, with manual adjustments as necessary. The annotations for other frames are generated mathematically using the two manually created annotations. The in-between frames still need to be validated to ensure accuracy, but we found that this only requires minor adjustments. We then export the annotations in a standard XML format (\textit{i.e.}, Pascal VOC)~\cite{Everingham15}. Once exported, we simply place annotations in the relevant folder created by the expansion tool.

Once the annotation process was completed for the initial data, the size of the dataset was too cumbersome for typical object detection methods and for ease of storage. At this point, the images included sizes that varied anywhere from $2k$ to $4k$ resolution, and a total dataset size of over $100$ gigabytes. To overcome this issue while producing an extensible dataset, we created a second command-line tool. This resizing tool targets all applicable folders within the target folder, so it can be utilized on a single annotation/image combo or on the entire dataset at once. Once the images and annotations are resized, the dataset storage size is drastically reduced (to about $5$ gigabytes in this case).

This resizing tool scales all properly formatted XML and image data to the specified size, whether smaller or larger than current size. We resize the images and annotations to $400 \times 400$ resolution to facilitate unified-size modeling, while the original images are also made available. The resize tool can optionally display all the images with their bounding boxes and labels. The images displayed in Figure~\ref{fig:dataset_example} are an example output of this tool. This simply allows images to be visually inspected as the dataset is resized. Some major annotation errors in famous datasets been found~\cite{northcutt2021pervasive}. This visual inspection process allows additional validation of annotations as FOD-A expands, which aids in the prevention of similar errors~\cite{northcutt2021pervasive}.

\subsection{Dataset Statistics}\label{dataset-stats}
After the initialization of the FOD-A dataset, there are a total of $31$ object categories and over 30,000 annotation instances. Figure~\ref{fig:annotation_instances} shows instances per category for the bounding box annotations and Table~\ref{tbl:categorization} shows the statistics for the light-level and weather categories.

\vspace{-1em}
\begin{table}[H]
\renewcommand{\arraystretch}{1.3}
\caption{Categorization Statistics}
\vspace{-.6em}
\label{tbl:categorization}
\centering
\begin{tabular}{| c c | c c c |}
\hline
\multicolumn{2}{|c|}{\textbf{Weather}} & \multicolumn{3}{|c|}{\textbf{Light-Level}}\\
\hline
Dry & Wet & Dark & Dim & Bright\\
\hline
26647 & 7216 & 4387 & 12464 & 17012\\
\hline
\end{tabular}
\vspace{-1em}
\end{table}

The material recognition dataset discussed in Section~\ref{sec:related}~\cite{FOD-conv-nn}, contained a total of $3$ object categories and $3,440$ annotation instances. As such, the potential applications of FOD-A and the materials recognition dataset~\cite{FOD-conv-nn} may differ greatly. A few datasets may be considered when analysing FOD-A, such as the Pascal VOC~\cite{Everingham15} and the Microsoft COCO~\cite{lin2014microsoft} datasets. However, the object categories contained in these datasets are of everyday objects and have a more general application scope than FOD-A. Image datasets of debris in airports should contain FOD specific object categories. For this reason, datasets of everyday objects are not directly comparable to FOD-A.

\section{Algorithmic Analysis}\label{sec:analysis}

\begin{figure*}[t]
    \centering
    \vspace{1mm}
    \includegraphics[width=400pt]{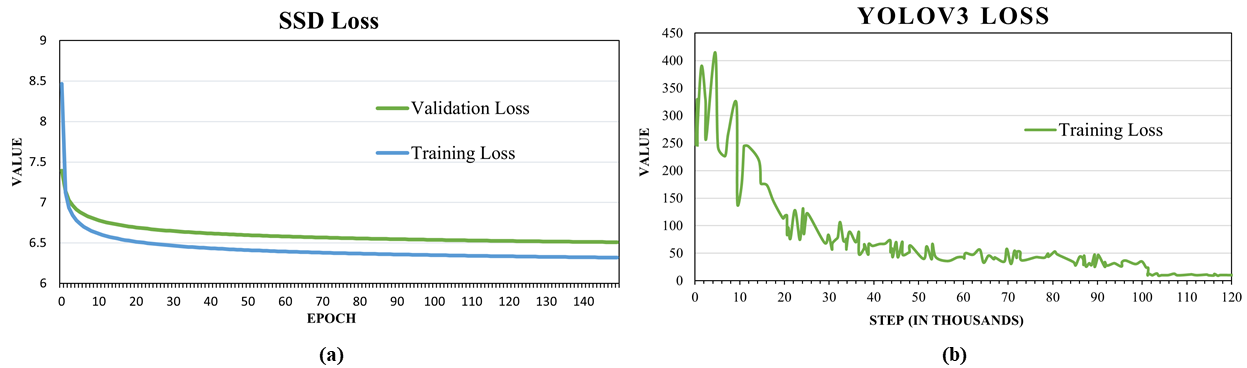}
    \vspace{-1.em}
    \caption{Loss curve from SSD and YOLO experiments.}
    \vspace{-1.5em}
    \label{fig:statistics}
\end{figure*}

To validate dataset functionality in abstract scenarios, it is necessary to implement FOD-A in several common MLCV algorithms. As mentioned previously, this dataset focuses on bounding-box based object detection functionality, but the potential of FOD-A is expanded by including categorization annotations. The categorization annotations are tested using a simple binary classification model described later in this section. The viability of the bounding box annotations are tested using the two famous algorithms: You Only Look Once Version 3 (YOLOv3)~\cite{redmon2018yolov3} and Single-Shot Multi-box detector (SSD)~\cite{DBLP:journals/corr/LiuAESR15}. Existing implementations of SSD~\cite{SSDImplementation} and YOLOv3~\cite{YOLOv3Implementation} are used. The categorization accuracy and mean intersection over union~\cite{girshick2014rich} (IOU) metrics are used to compare results between the algorithms. Both metrics are calculated using predictions on the validation dataset. Accuracy is computed as the number of correct categorization predictions over the total number of predictions. To calculate the mean IOU~\cite{girshick2014rich}, the IOU value of predictions in true positive categorization cases are averaged. IOU values from only true positive categorization cases facilitates separation from accuracy results.

The initial experimentation began using YOLOv3. As shown in Figure \ref{fig:statistics}(b), the loss reaches a value of about $7.05$. For this implementation~\cite{YOLOv3Implementation}, the loss is calculated using the method presented in the original YOLOv3 paper~\cite{redmon2018yolov3}. In this experimentation, YOLOv3 produces categorization accuracy of $12.42\%$ and a mean IOU of $47.58\%$ on FOD-A validation data. As the accuracy and mean IOU metrics suggest, the YOLOv3 algorithm commonly predicts an incorrect categorization label, but regularly produces correct bounding boxes.

As shown in Figure~\ref{fig:statistics}(a), SSD loss reaches a value of about 6.51; the loss result approaches convergence after about 140 epochs. The loss in this implementation~\cite{SSDImplementation} is calculated using the method defined in the original SSD paper~\cite{DBLP:journals/corr/LiuAESR15}. SSD provides categorization accuracy of $71.81\%$ and a mean IOU of $68.05\%$.

In this experimentation, SSD produced better results than YOLOv3. However, the scope of this paper is mostly restricted to the presentation of the FOD-A dataset, as algorithm optimization for FOD-A is future work. As intended, FOD-A proved to be difficult for the modern YOLOv3 and SSD algorithms. This provides room for future algorithmic enhancement, in both efficiency and accuracy.

Since FOD-A also includes categorization annotations, we examine this functionality with a binary categorization model using transfer learning. With the output layer removed and substituted for both a max pooling and a fully-connected layer with two output neurons, this binary classification model was built using the MobileNetv2~\cite{sandler2019mobilenetv2} architecture with weights pretrained on ImageNet~\cite{5206848}. FOD-A includes weather and light categorization annotations. To test the functionality of the categorization annotations, we conducted experiments using the two weather annotations, \textit{wet} and \textit{dry}. The binary categorization model quickly became skillful, and was able to effectively distinguish between a \textit{wet} and \textit{dry} background in FOD-A images. The accuracy quickly improved to the maximum percentage on validation data. The model is capable of correctly categorizing most images, but there were still some outlier predictions on the testing data. Although the weather classification alone can be solved quickly, the combination of the categorization and bounding box detection prove difficult for modern algorithms. Additionally, the weather and light annotations could prove useful in future practical work.

\section{Conclusion}\label{sec:conclusion_discussion}
MLCV has produced promising results for various tasks in FOD. However, to the best of our knowledge, a proper publicly available dataset of FOD has not been initialized prior to the proposed work. This paper introduces FOD-A and proposes an abstractable method of image dataset creation. As discussed previously, FOD-A object categories have been selected based on prior FAA documentation and research. This enables comprehensive coverage of common FOD. Moreover, we have developed an efficient and publicly documented expansion process and intend to make several extensions of FOD-A available on the GitHub repository. An efficient expansion process is important since FOD is a continually evolving datatype. To simulate airport environments, images in FOD-A contain varying light and weather conditions. In addition to the bounding box annotation, we provide these weather and light conditions as categorization labels. We also validate these approaches for both the object detection and the categorization functionalities of FOD-A on different algorithms. The experimental results demonstrate FOD-A's practicality and difficulty.

There are several research paths that could be followed to build on this work. One direction for future work is to develop more efficient and accurate object detection techniques for the FOD datatype. Once detection algorithms are improved, further works could explore practical experimentation in airports.

\section{Resources}
\label{resources}

GitHub: \href{https://github.com/FOD-UNOmaha/FOD-data}{https://github.com/FOD-UNOmaha/FOD-data}

\bibliographystyle{IEEEtran}
\bibliography{IEEEabrv,references}


\end{document}